\documentclass{article}

\PassOptionsToPackage{numbers}{natbib}
\usepackage[final]{tackling_climate_workshop_style}

\usepackage[utf8]{inputenc} 
\usepackage[T1]{fontenc}    
\usepackage{hyperref}       
\usepackage{url}            
\usepackage{booktabs}       
\usepackage{amsfonts}       
\usepackage{nicefrac}       
\usepackage{microtype}      

\usepackage{graphicx}
\usepackage{subcaption}
\usepackage{appendix}
\usepackage{tikz}
\usetikzlibrary{trees}
\title{Hierarchical Classification for Automated Image Annotation of Coral Reef Benthic Structures}

\author{%
  Célia Blondin$^1$, Joris Guérin$^1$, Kelly Inagaki$^2$, Guilherme Longo$^2$ \& Laure Berti-Équille$^1$ \\
  $^1$ Espace-Dev, IRD, Univ. Montpellier, Montpellier, France\\
  $^2$ Universidade Federal do Rio Grande do Norte, Natal, Brazil\\
  Correspondance: \texttt{celia.blondin@ird.fr}, \texttt{joris.guerin@ird.fr} \\
}

\begin{document}

\maketitle

\vspace{-10pt}

\begin{abstract}
  Automated benthic image annotation is crucial to efficiently monitor and protect coral reefs against climate change. Current machine learning approaches fail to capture the hierarchical nature of benthic organisms covering reef substrata, i.e., coral taxonomic levels and health condition. To address this limitation, we propose to annotate benthic images using hierarchical classification. Experiments on a custom dataset from a Northeast Brazilian coral reef show that our approach outperforms flat classifiers, improving both F1 and hierarchical F1 scores by approximately 2\% across varying amounts of training data. In addition, this hierarchical method aligns more closely with ecological objectives.
\end{abstract}

\vspace{-10pt}

\section{Introduction}

Coral reefs host around 25\% of global marine species~\cite{souter_status_2020} and provide vital ecosystem services for coastal populations, including maintaining fish stocks, protecting shorelines from waves, and fostering tourism~\cite{woodhead2019coral}. Despite their importance, these ecosystems are facing unprecedented threats from climate change~\cite{spalding2015warm}
as illustrated by the increasing frequency of coral bleaching events~\cite{eakin_20142017_2019}. In response to these challenges, comprehensive coral reef surveying and monitoring have become essential for implementing appropriate and timely conservation actions~\cite{edgar2014systematic}.

\begin{figure}[t]
    \centering
    \includegraphics[width=0.8\textwidth]{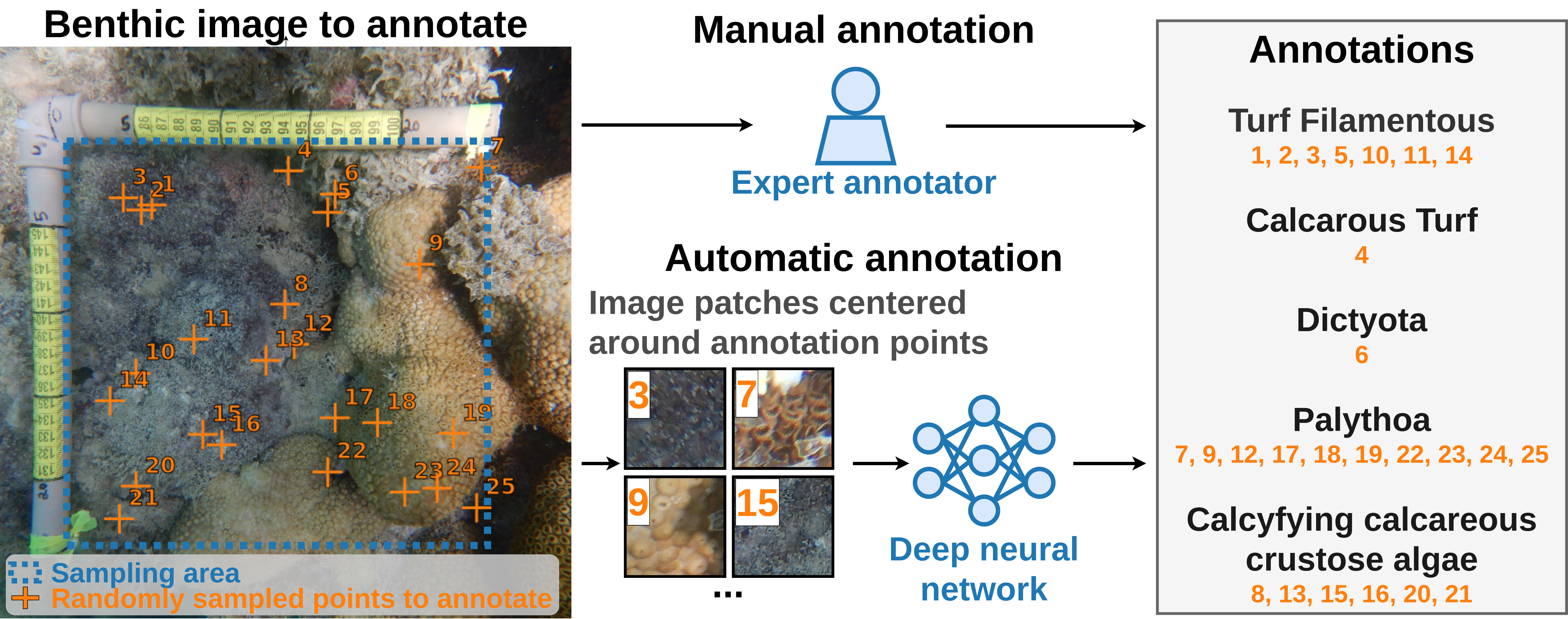}
    \caption{\textbf{Benthic image annotation process} -- A predetermined number of random points 
    are sampled within a fixed area and 
    annotated with the corresponding organism or substrate. The relative frequencies are used as proxies for cover proportions of benthic categories within the sampled area.}
    \label{fig:annotation_example}
\end{figure}

Benthic monitoring focuses on measuring seafloor coverage by various organisms, especially corals~\cite{de201227}. Over the past decades, monitoring programs have shifted from direct underwater measurements to photographic methods
~\cite{jokiel2015comparison}, creating lasting records and significantly reducing underwater time. Traditionally, collected images were annotated manually using specialized software,
involving sampling random points and labeling them with their benthic categories (Figure~\ref{fig:annotation_example}). These annotations serve to calculate cover proportions, which are essential for coral monitoring.

However, manual annotation is time-consuming, which can delay the availability of survey results, and can introduce bias due to differences between analysts~\cite{beijbom2015towards}. To address this, platforms like CoralNet~\cite{williams2019coralnet} use deep learning for automatic annotation~\cite{chen2021coralnet2}. Small pixel patches (typically 224x224) are extracted around each point and the problem of point annotation is addressed as an image classification task (Figure~\ref{fig:annotation_example}). This way, ecologists only need to manually annotate a small subset of images, which are used to train a neural network capable of automatically annotating the remaining images.

During our collaboration between data scientists and marine ecologists, 
we identified a misalignment between Machine Learning (ML) approaches and ecological needs. Current models use detailed labels encompassing various taxonomic levels and coral health status. For instance, a point labeled as ``Bleached Palythoa" simultaneously conveys information about coral type, genus, and health status (Figure~\ref{fig:tree}).
However, in practice, ecologists often need to aggregate these labels differently to compute cover proportions of macro groups, e.g., grouping all corals regardless of species or health.

Flat classification presents challenges for underrepresented categories, which often lack sufficient labeled data for accurate classification at such a high level of detail. To address this issue, we investigate the application of Hierarchical Classification (HC)~\cite{silla2011hierarchical_survey} to benthic image annotation, a novel approach in this field. This multi-level prediction strategy has the potential to improve performance by allowing partially correct predictions. Furthermore, we argue that evaluating model performance with hierarchical metrics~\cite{kosmopoulos2015hierarchical_metrics} would more accurately reflect real-world ecological practices. By aligning our approach with the inherent hierarchical structure of benthic organisms, we aim to enhance both the performance and ecological relevance of automated coral reef monitoring.

\section{Methodology}

\begin{figure}[t]
    \centering
    \includegraphics[width=0.8\textwidth]{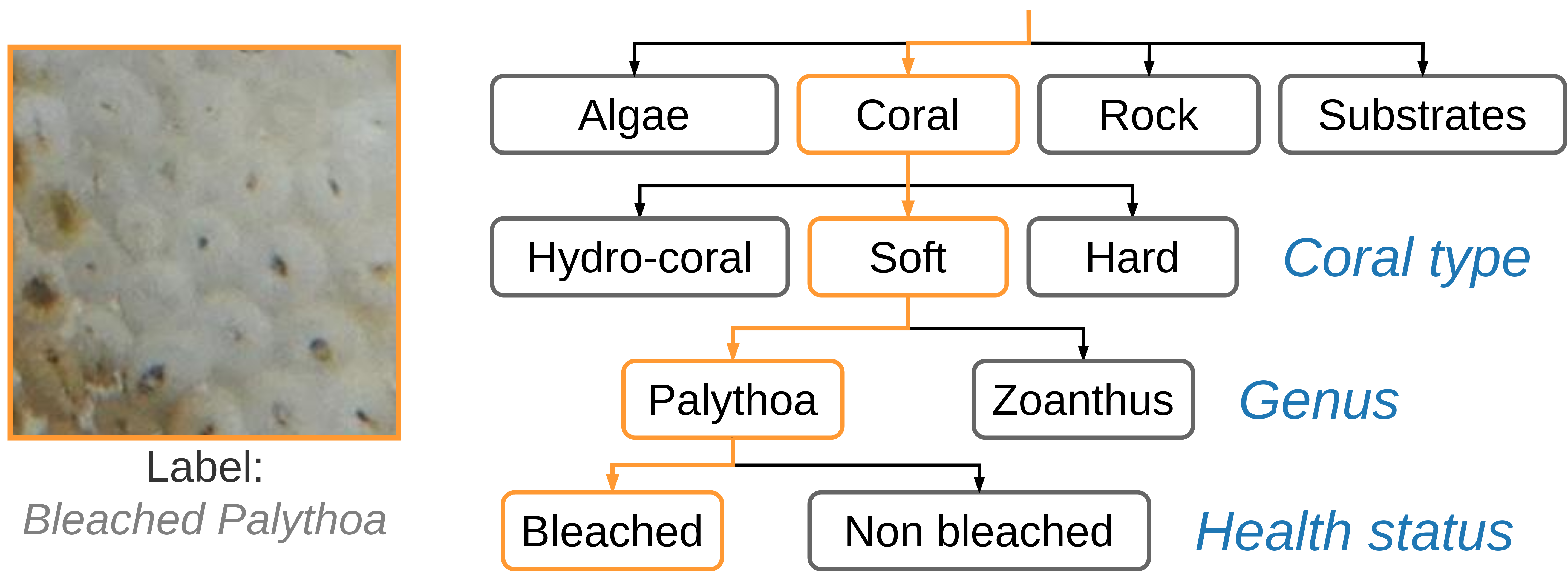}
    \caption{\textbf{Hierarchical structure of benthic labels} -- This conceptual diagram illustrates the multi-level categories used for classifying benthic organisms and substrates in coral reef ecosystems.}
    \label{fig:tree}
\end{figure}

Hierarchical Classification (HC) consists in organizing classes into a hierarchy, represented as a tree~\cite{wehrmann2018hierarchical, silla2011hierarchical_survey} (Figure~\ref{fig:tree}). This structure reflects the natural relationships between classes, with general categories at higher levels and specific subcategories at lower levels. Two main approaches exist: local approaches, often called ``top-down", train separate classifiers for each node of the hierarchy, while global approaches build a single, comprehensive model considering the entire hierarchy simultaneously. In this paper, we evaluate the potential of top-down HC for benthic image annotation.

\subsection{Building the Hierachy}Implementing HC for coral images requires a paradigm shift in how the label set is designed. Marine ecologists must establish a classification tree representing the taxonomic hierarchy, rather than a flat set of labels. 
Building an HC tree requires answering several design questions: Can a prediction be composed of multiple labels? Can multiple path lead to the same label? Can predictions stop at intermediate nodes? As the goal of this work is to validate the relevance of HC for benthic image annotation, we choose the most simple tree design possible where the final prediction is always a single leaf node, and there is only one path to each label. 
The tree (Appendix~\ref{appendix:tree}) was designed in a collaboration between an ML expert and a marine ecologist to ensure both computational constraints and biological relevance are respected. 
While this initial step represents an additional effort, it helps aligning the annotation process with biological classification principles and facilitates subsequent ecological analyses involving grouping labels at various taxonomic levels. 

\subsection{Hierarchical Classifier}
The first step to annotate an image is to extract 224x224 patches around the annotation points. Each patch is passed through the CoralNet backbone, an EfficientNet B0 pre-trained on 16 million benthic images~\cite{chen2021coralnet2}, to extract features. 
Then, a top-down hierarchical classifier is implemented. We use the hiclass library~\cite{miranda_hiclass_2023} and train a \textit{Local Classifier per Parent Node} approach. This approach consist of using an independent classifier for each non-leaf node. In our case, we use MLPs with two hidden layers (200 and 100 neurons) as the individual node classifiers.

\section{Experiments}
This section presents the experiments conducted to assess the potential of HC for benthic image annotation. We provide a detailed Github repository to reproduce our results\footnote{\url{https://github.com/celia-bl/hierarchical_classifying_corals_dataset.git}}.
\subsection{Experimental Setup}
\subsubsection{Dataset} For the experiments, we used a custom dataset, collected and annotated by the Marine Ecology Laboratory/UFRN, at Rio do Fogo, Northeast Brazil. Between 2017 and 2022, 100 images were collected from 2 sites every 3 months. 
We utilized a subset of 1,549 images, 
each annotated with 25 random points, resulting in 38,725 annotations. The dataset contains 54 distinct labels, which are highly imbalanced -- 11 labels account for 95\% of the annotations, while 24 classes count less than 20 annotations. The distribution of annotate patches is detailed in Appendix~\ref{appendix:class_distribution}.
We created a test set by randomly selecting 10\% of the dataset, maintaining the overall class distribution. Test patches were chosen independently, without considering image-level coherence.

\subsubsection{Baseline} 
Current approaches typically train a flat classifier with a large set of labels corresponding to all leaf nodes~\cite{chen2021coralnet2}. Ecologists can subsequently group these labels to calculate the cover of super-classes, such as ``all corals"~\cite{aued2018large}. This post-hoc grouping task can be viewed as ``bottom-up" hierarchical classification. In this work, we compare a traditional bottom-up approach -- MLP with 2 hidden layers (200, 100) -- against the proposed top-down approach, which considers the hierarchy during training.

\subsubsection{Evaluation Metrics} 
Traditional benthic image annotation approaches employ standard classification metrics (e.g., accuracy, precision, recall, F1-score), evaluating a model's ability to make perfect predictions across all taxonomic levels. However, considering the hierarchical nature of benthic classifications, not all misclassifications have equal ecological significance. For instance, misclassifying a Bleached Palythoa as a non-bleached Palythoa has less impact on derived ecological metrics than misclassifying it as Algae. The former error affects only leaf-node accuracy, while the latter affects cover estimates at all levels. To address this nuance, we use Hierarchical Classification (HC) metrics~\cite{kosmopoulos2015hierarchical_metrics}, which account for the varying severity of classification errors based on their position in the hierarchy. In our study, we compare top-down and bottom-up approaches using both the traditional F1-score (F1) and the hierarchical F1-score (hF1). This dual evaluation allows us to assess the impact of incorporating hierarchical information during training on both perfect label prediction and partially accurate predictions. By doing so, we aim to provide a more comprehensive evaluation that aligns with the hierarchical nature of benthic classification and its ecological implications.

\subsection{Results}

\begin{figure}[t]
    \centering
    \begin{subfigure}[b]{0.45\textwidth} 
        \centering
        \includegraphics[width=\textwidth]{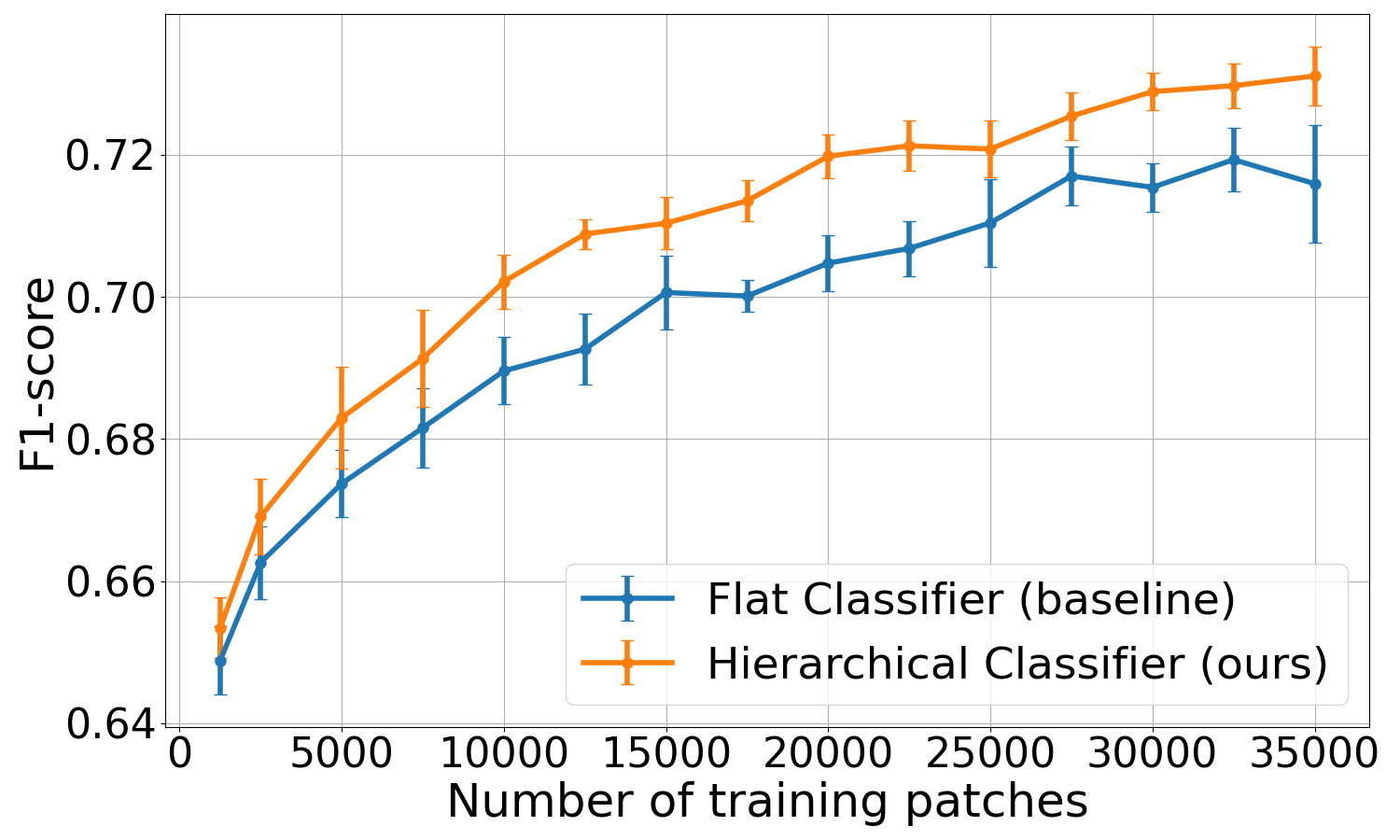} 
        \caption{F1-score}
        \label{fig:f1_plot}
    \end{subfigure}
    \hspace{0.03\textwidth} 
    \begin{subfigure}[b]{0.45\textwidth} 
        \centering
        \includegraphics[width=\textwidth]{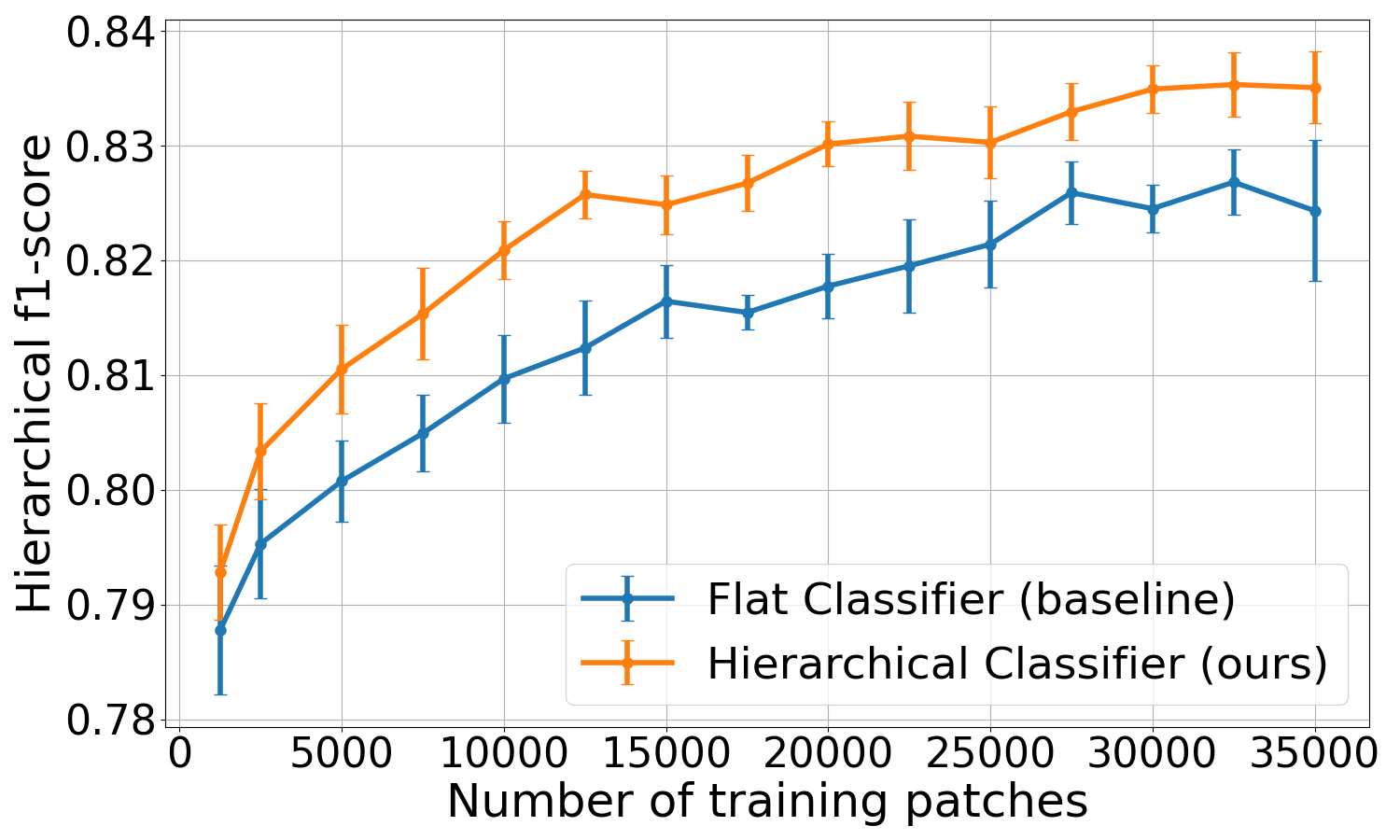} 
        \caption{Hierarchical F1-score}
        \label{fig:hierarchical_f1_plot}
    \end{subfigure}
    \vspace{4pt}
    \caption{\textbf{Experimental results} -- Comparison of flat and hierarchical classifiers for our custom benthic image annotation dataset, showing performance metrics across varying training data sizes. Error bars indicate standard deviation from random training set sampling.}
    \label{fig:comparison_TD_BU}
\end{figure}

Our experimental results are displayed in Figure~\ref{fig:comparison_TD_BU}, where we compare the F1-score and hierarchical F1-score for the flat and hierarchical classifiers when the number of training patches increases. It shows that the hierarchical classifier always outperforms the flat classifier, with performance gains ranging from 1\% (for small training sets) to 2\% (for larger sets).

Interestingly, the performance gap between the two approaches remains similar for the hierarchical F1-score, contrary to our expectation that this metric would more severely penalize errors early in the tree structure for the flat classifier. This suggests that the flat classifier may implicitly capture some of the hierarchical relationships between classes. If the flat classifier were making largely implausible errors, we would expect a more substantial difference in the hierarchical F1-score.

\section{Conclusion}

In this work, we investigated the potential benefits of using hierarchical classification for benthic image annotation. Our experimental results indicate that HC outperforms traditional flat classifiers across both flat and hierarchical metrics. 
However, the flat classifier is more robust than anticipated to hierarchical F1-score, possibly due to its ability to learn some inherent class relationships. Further investigation into the specific types of errors made by each classifier could provide additional insights into their respective strengths and limitations.

To gain a deeper understanding of the performance of HC for benthic image annotation, we plan to apply this methodology to other datasets to contextualize these results and assess their significance. For instance, the Moorea Labeled Corals dataset~\cite{moorea_coral_reef_lter_mcr_2019} contains a larger number of annotations and a simpler hierarchical structure, which could provide additional insights.

Although building the tree represents an additional effort, it enables better consideration of the labels used, avoids overlapping categories or unnecessary labels, and facilitates subsequent ecological analyses, which often involve grouping labels at various taxonomic levels. To alleviate the workload of tree construction, a promising research direction to improve our approach is to investigate how this process can be automated and refined to enhance class discrimination~\cite{srivastava_discriminative_2013}. Additionally, exploring alternative hierarchical classification methods, such as global HC, could represent a valuable contribution.

Finally, while our results demonstrate the potential of HC, the modest performance gains (~2\%) necessitate careful consideration of the trade-off between improved accuracy and increased computational costs associated with training and inferring with multiple classifiers in a hierarchical setups. This balance will be crucial in determining the applicability of HC in benthic image annotation tasks.

\newpage
\small
\bibliographystyle{IEEEtranN}

\appendix 

\section{Hierarchical Tree} \label{appendix:tree}
\resizebox{!}{0.95\textheight}{
\begin{tikzpicture}[
  >=stealth, 
  every node/.style={rectangle, draw, rounded corners, text centered, font=\small}, grow=right,
  level distance=2cm, edge from parent path={(\tikzparentnode) -- (\tikzchildnode)}
]

\node {Root}[level distance=2.5cm]
  child { node {Algae}[sibling distance=3cm, level distance=2cm]
    child { node {Non Calcified}[sibling distance=0.7cm, level distance=6cm]
      child { node {Caulerpa}[sibling distance=0.6cm, level distance=3cm]
        child { node {Caulerpa} } 
        child { node {Caulerpa} } 
        child { node {Caulerpa Cupressoides} } 
      }
      child { node {Pterocladiella} } 
      child { node {Gelidiaceae}[sibling distance=0.6cm, level distance=3cm]
        child { node {Gelidium spp.} } 
        child { node {Gelidiella spp.} } 
      }  
      child { node {Wrangelia} } 
      child { node {Cladophora spp.} } 
      child { node {Dictyopteris}[sibling distance=0.6cm, level distance=3.5cm]
        child { node {Dictyota Mertensii} } 
        child { node {Dictyopteris spp.} } 
        child { node {Dictyopteris Plagiogramma} } 
        child { node {Dictyopteris Delicatula} } 
        child { node {Dictyopteris Jamaicensis} } 
      }
      child[shift={(-0.1,0)}] { node {Macroalgae: Filamentous} } 
      child { node {Hypnea Musciformis} } 
      child[shift={(-0.2,0)}] { node {Leathery Macrophytes: Other} } 
      child { node {Codium} } 
      child { node {Dictyota}[sibling distance=0.6cm, level distance=3cm]
        child { node {Dictyota Ciliolata} } 
        child { node {Dictyota Menstrualis} } 
        child { node {Dictyota spp.} } 
        child { node {Dictyota spp.} } 
        child { node {Canistrocarpus Cervicornis} } 
      }
      child { node {Sargassum} } 
      child { node {Valonia Ventricosa} } 
      child { node {Laurencia spp.} } 
      child { node {Turf}[sibling distance=0.6cm, level distance=3cm]
        child { node {Turf Filamenteous} } 
        child { node {Turf and sand} } 
        child { node {Calcareous Turf} } 
      }
    }
    child[shift={(-0.7,0.4)}] { node {Calcified} [sibling distance=0.6cm, level distance=3cm]          child[shift={(-1,0.4)}] { node {Amphiroa spp.} } 
child[shift={(0,0.6)}] { node {Macroalgae:\newline Articulated calcareous} } 
child[shift={(-0.5,0.9)}] { node {Calcifying calcareous crustose algae: DHC} } 
    }
  }
  child { node {Fish} } 
  child { node {Rock} } 
  child[shift={(0.3,0.3)}] { node {Other invertebrates}[sibling distance=0.7cm, level distance=2.5cm]
    child { node {Ascidian} } 
    child { node {Anemone} } 
    child { node {Sponges}[sibling distance=0.7cm, level distance=2cm]
      child { node {Sponge} } 
      child { node {Placospongia} } 
      child { node {Encrusting sponge} } 
    }
  }
  child { node {Unknown} } 
  child { node {Cyanobacteria}[sibling distance=0.7cm, level distance=3.2cm]
    child { node {Cyanobacteria films} } 
    child { node {Cyanobacteria} } 
  }
  child { node {Substrates}[sibling distance=0.7cm, level distance=2cm]
    child[shift={(0.3,0)}] { node {Unconsolidated (soft)} } 
    child { node {Sand} } 
  }
  child { node {Shadow} } 
  child { node {Corals}[sibling distance=2.3cm, level distance=3.5cm]
    child[shift={(1,0.5)}] { node {Soft}[sibling distance=1.9cm, level distance=2cm]
      child[shift={(0,-0.5)}] { node {Zoanthidae}[sibling distance=1.4cm]
        child { node {Palythoa spp.}[sibling distance=0.6cm, level distance=3cm]
          child { node {Palythoa Caribaeorum} } 
          child { node {Palythoa spp.} } 
          child { node {Protopalythoa Variabilis} } 
        }
        child { node {Zoanthus spp.}[sibling distance=0.6cm, level distance=3cm]
          child { node {Zoanthus spp.} } 
          child { node {Zoanthus Sociatus} } 
        }
      }
      child[shift={(0,-0.3)}] { node {Octocoral}
        child { node {Plexauridae}[sibling distance=0.6cm, level distance=3cm]
          child { node {Plexaurella Grandiflora} } 
          child { node {Plexaura spp.} } 
        }
      }
    }
    child[shift={(0,-0.1)}] { node {Hydro-corals}[level distance=4cm]
      child { node {Millepora}[sibling distance=0.6cm, level distance=3cm]
        child[shift={(0,0.1)}] { node {Millepora spp.} } 
        child { node {Millepora Alcicornis} } 
      }
    }
    child[shift={(-1,-0.6)}] { node {Bleached}[sibling distance=0.6cm, level distance=2.3cm]
      child { node {Bleached Hard Coral} } 
      child[shift={(0.4,0)}] { node {Soft Coral Bleached} } 
      child[shift={(0.3,0)}] { node {Bleached Coral Point} } 
      child[shift={(0.2,0)}] { node {Dead Coral} } 
      child { node {Recent Dead Coral} } 
    }
    child { node {Hard}[sibling distance=0.8cm, level distance=4cm]
      child { node {Favia}[sibling distance=0.6cm, level distance=3cm]
        child { node {Favia Gravida} } 
        child { node {Favia Leptophylla} } 
      }
      child { node {Porites}[level distance=3cm]
        child { node {Porites Astreoides} } 
      }
      child { node {Siderastrea}[sibling distance=0.6cm, level distance=3cm]
        child { node {Siderastrea spp.} } 
        child { node {Siderastrea Stellata} } 
      }
      child { node {Tubastrea}[level distance=3cm]
        child { node {Tubastrea} } 
      }
      child { node {Mussimila}[level distance=3cm]
        child { node {Mussimila} } 
      }
      child { node {Agaricia}[sibling distance=0.6cm, level distance=3cm]
        child { node {Agaricia Fragilis} } 
        child { node {Agaricia Humilis} } 
      }
    }
  };
\end{tikzpicture}}

\section{Class Distribution}\label{appendix:class_distribution}
\begin{figure}[h]
    \centering
    \includegraphics[width=\textwidth]{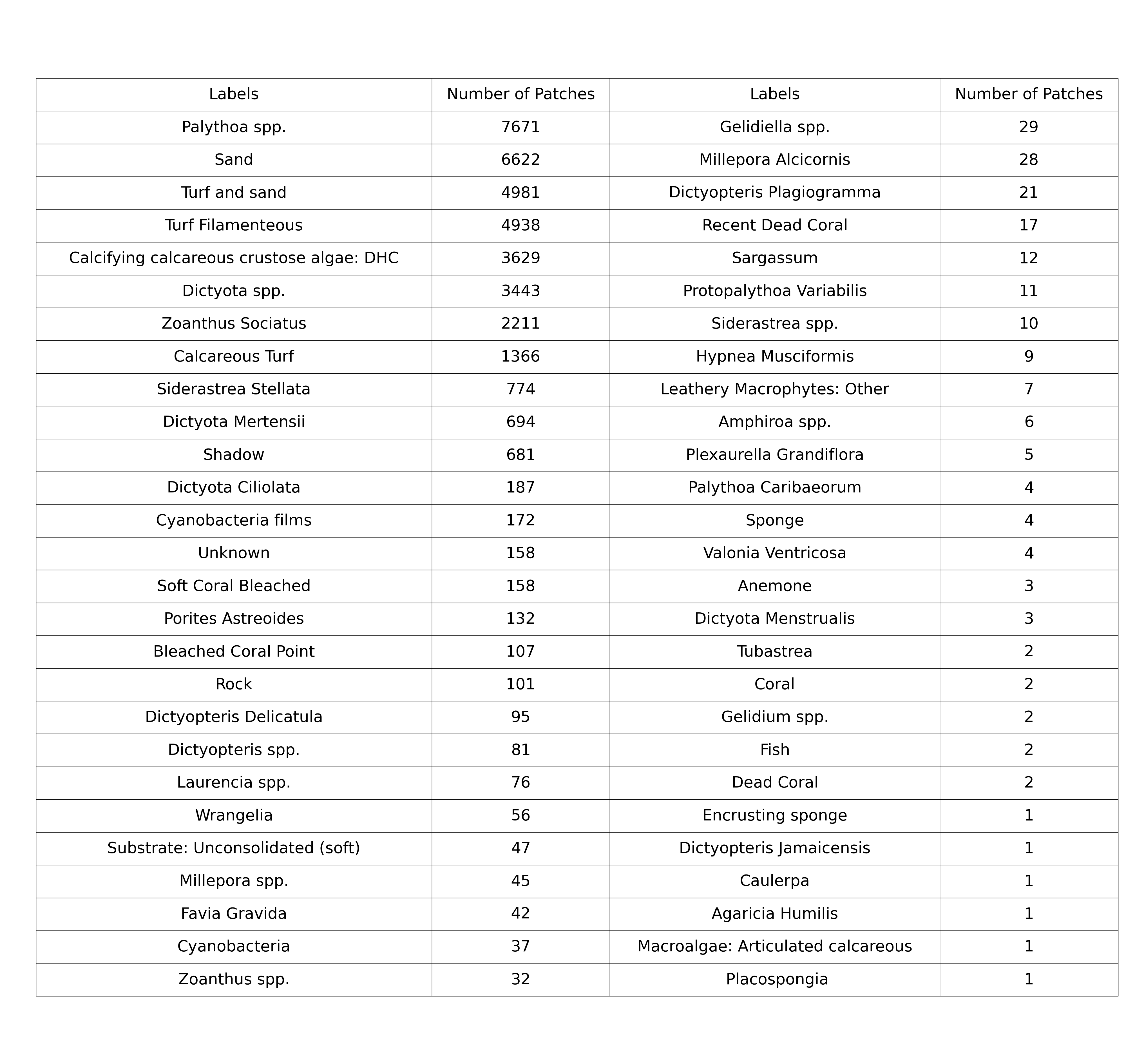}
    \caption{Number of patches corresponding to each leaf node present in our custom benthic image annotation dataset.}
    \label{fig:complete_tree}
\end{figure}

\end{document}